\pdfoutput=1

\documentclass[11pt]{article}

\usepackage[final]{coling}

\usepackage{times}
\usepackage{latexsym}

\usepackage[T1]{fontenc}

\usepackage[utf8]{inputenc}

\usepackage{microtype}

\usepackage{inconsolata}

\usepackage{graphicx}

\usepackage{subcaption}
\usepackage{booktabs}
\usepackage{multirow}

\usepackage{CJKutf8}

\usepackage{amsmath}

\usepackage{algorithm}
\usepackage{algorithmic}

\title{Refining Dimensions for Improving Clustering-based \\Cross-lingual Topic Models}

\author{
 \textbf{Chia-Hsuan Chang\textsuperscript{1}},
 \textbf{Tien-Yuan Huang\textsuperscript{2}},
 \textbf{Yi-Hang Tsai\textsuperscript{2}},
 \textbf{Chia-Ming Chang\textsuperscript{2}},
 \textbf{San-Yih Hwang\textsuperscript{2}}
\\
 \textsuperscript{1}Department of Biomedical Informatics \& Data Science, Yale University
\\
New Haven, CT 06510, United States
\\
 \textsuperscript{2}Department of Information Management, National Sun Yat-sen University
\\
Kaohsiung 80424, Taiwan
\\
 \small{
   \textbf{Correspondence:} \href{mailto:shane.chang.tw@gmail.com}{shane.chang.tw@gmail.com}, \href{mailto:syhwang@mis.nsysu.edu.tw}{syhwang@mis.nsysu.edu.tw}
 }
}

\begin{document}
\maketitle
\begin{abstract}
Recent works in clustering-based topic models perform well in monolingual topic identification by introducing a pipeline to cluster the contextualized representations. However, the pipeline is suboptimal in identifying topics across languages due to the presence of language-dependent dimensions (LDDs) generated by multilingual language models. To address this issue, we introduce a novel, SVD-based dimension refinement component into the pipeline of the clustering-based topic model. This component effectively neutralizes the negative impact of LDDs, enabling the model to accurately identify topics across languages. Our experiments on three datasets demonstrate that the updated pipeline with the dimension refinement component generally outperforms other state-of-the-art cross-lingual topic models~\footnote{Our code and data are available at \url{https://github.com/Text-Analytics-and-Retrieval/Clustering-based-Cross-Lingual-Topic-Model}.}.
\end{abstract}

\begin{CJK*}{UTF8}{bsmi}

\section{Introduction}

Traditional cross-lingual topic models~(CLTM) rely on additional resources to identify topics across languages. Based on the types of resources, CLTMs can be categorized into document and vocabulary-linking models. The document-linking models require parallel or comparable corpora to model the co-occurring word statistics across languages and infer the cross-lingual topics~\citep{mimnoPolylingualTopicModels2009,piccardiCrosslingualTopicModeling2021}. The vocabulary-linking models are more resource-efficient than their document-linking counterpart because they only require a bilingual dictionary (i.e., a set of translation entries). However, vocabulary-linking models often result in monolingual topics~\citep{huPolylingualTreeBasedTopic2014,haoEmpiricalStudyCrosslingual2020,wuInfoCTMMutualInformation2023} when the dictionary is of limited coverage to the target corpus. Several studies proposed to link word embedding spaces across languages to decrease the effort of compiling a well-covered dictionary. When the assumption of shared structures across spaces (i.e., isomorphism) holds, a small number of translation entries will be sufficient to identify topics across languages~\citep{changIncorporatingWordEmbedding2018,yuanMultilingualAnchoringInteractive2018,changWordEmbeddingbasedApproach2021}. However, the word spaces of different languages seldom share the same structure in practice, especially for languages that are distantly related, and iterative human involvement is still required for acquiring a quality dictionary.

\begin{figure}[ht]
	\centering
	\begin{subfigure}[l]{0.45\linewidth}
		\centering
            \includegraphics[width=.6\columnwidth]{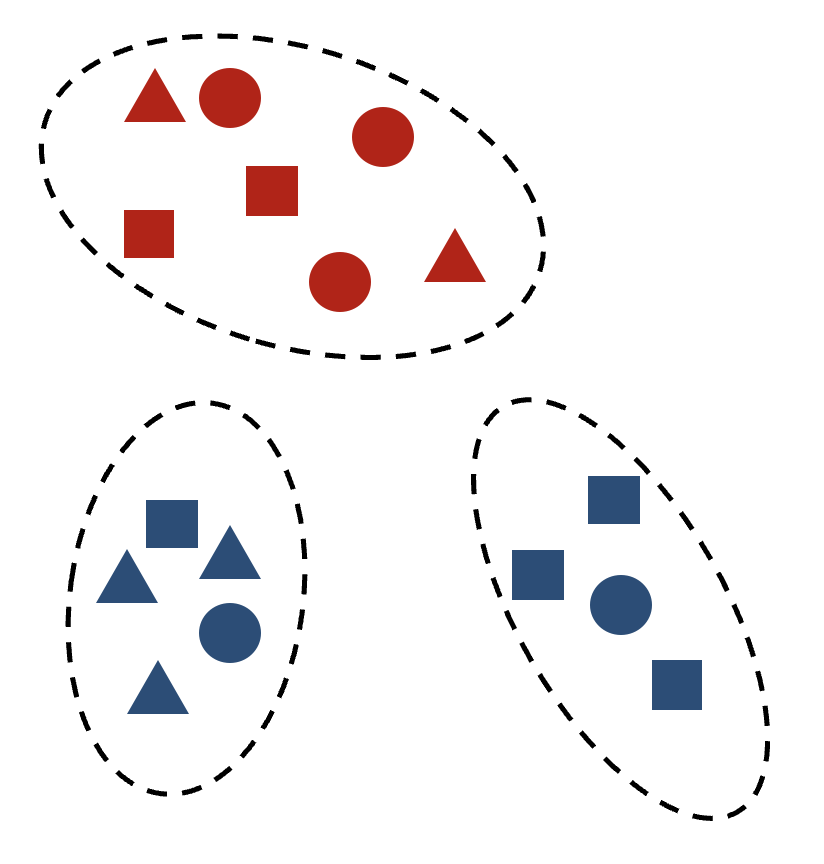}  
		\subcaption{Topic clusters grouped by languages}
            \label{fig:clustering_a}
	\end{subfigure}
	\begin{subfigure}[r]{0.45\linewidth}
		\centering
            \includegraphics[width=.6\columnwidth]{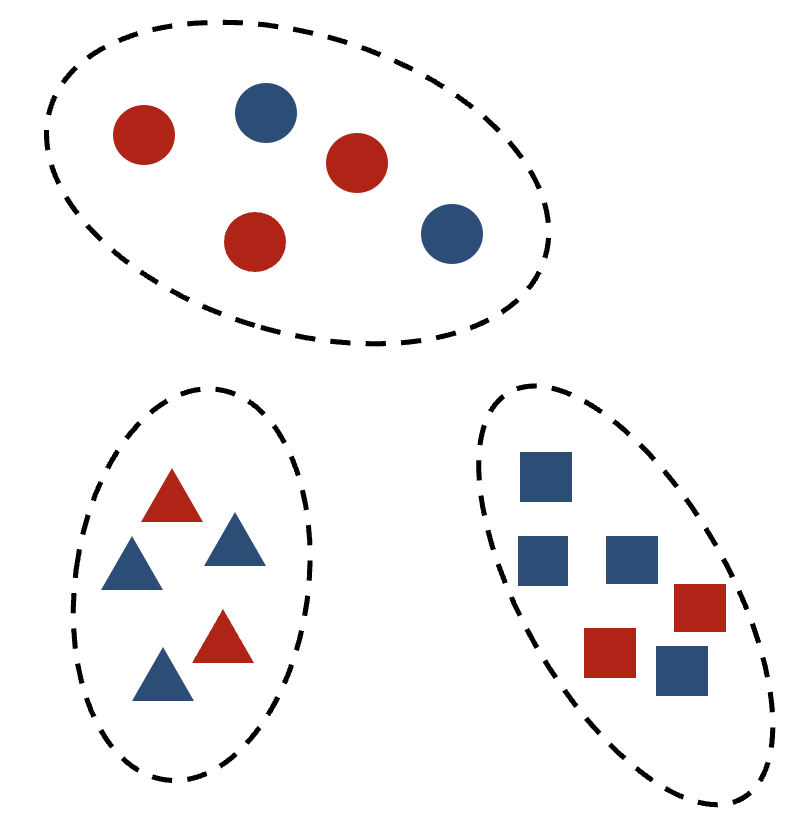}  
		\subcaption{Topic clusters grouped by semantic meanings}
            \label{fig:clustering_b}
	\end{subfigure}
    \label{fig:clustering}
    \caption{Two resultant scenarios of clustering-based topic model. Different shapes indicate the documents discussing various topics, while different colors represent documents of different languages.}
\end{figure}

The recent development of multilingual language models (MLM), e.g., mBERT, XLM-R, and GPT models, attracts attention from the natural language processing community. MLM learns the language-agnostic representations without any additional resources~\cite{piresHowMultilingualMultilingual2019,dufterIdentifyingElementsEssential2020}, which has the potential to realize the zero-shot topic identification across languages~\cite{bianchiCrosslingualContextualizedTopic2021}, thereby reducing efforts on data preparation. Recent studies increasingly favor the clustering-based topic model due to its superior performance and higher efficiency~\cite{siaTiredTopicModels2020,grootendorstBERTopicNeuralTopic2022,zhangNeuralTopicModelling2022}. The clustering-based topic model adopts a pipeline (see Sec.~\ref{sec: pipeline of clustering-based topic model}) to leverage the induced representations of language models for topic identification. MLMs can be directly applied to the pipeline of clustering-based topic modeling for cross-lingual topic identification. However, the current pipeline is hindered by the existence of language-dependent dimensions (LDDs) in the representations generated by MLMs, which makes the representations sensitive to languages and hinders the pipeline from identifying topics across languages. As depicted in Fig.~\ref{fig:clustering_a}, the current pipeline with MLM tends to cluster documents by languages rather than semantic meanings. We also report the qualitative result of misaligned topics generated using BERTopic~\citep{grootendorstBERTopicNeuralTopic2022}, an accessible implementation for clustering-based topic modeling, in Table~\ref{tbl:topic_words}. Ideally, topic clusters should group documents based on their semantic meanings, as illustrated in Fig.~\ref{fig:clustering_b}.

\begin{table*}[ht]
\caption{Top representative words of five sampled topics generated from BERTopic~\citep{grootendorstBERTopicNeuralTopic2022} with default parameters. We first use Cohere multilingual model to embed the Airiti dataset~\citep{changEmployingWordMover2020} and then employ BERTopic to generate topics.}
\label{tbl:topic_words}
\centering
\resizebox{0.85\textwidth}{!}{
  \begin{tabular}{ll}
    \midrule
    Topic\#1& cell, protein, expression, induce, gene, mouse, find, show, study, treatment\\
    Topic\#2& 細胞(cell), 蛋白(protein), 表現(expression), 基因(gene), 抑制(inhibition)\\
     & 蛋白質(protein), 我們(we), 發現(discover), 調控(control), 病毒(virus)\\
    Topic\#5& firm, market, financial, company, return, investor, investment, bank, stock, model\\
    Topic\#22& 反應(reaction), 分子(molecule), 高分子(polymer), 結構(structure), 合成(synthesize)\\
     & 化合物(compound), 錯合物(complex), 具有(have), 形成(form), 利用(utilize)\\
    Topic\#46& 市場(market), 報酬(return), 投資(investment), 股票(stock), 指數(index)\\
     & 股價(stock price), 交易(transaction), 模型(model), 公司(company), 價格(price)\\
  \bottomrule
\end{tabular}
}
\end{table*}

To mitigate such a problem, this study proposes adding a new dimension refinement component into the pipeline to neutralize the impacts of LDDs from the representations. Specifically, we utilize singular value decomposition (SVD) to identify the LDDs and offer two implementations of the dimension refining component: unscaled SVD (u-SVD) and SVD with language dimension removal (SVD-LR). The contributions of this study are threefold:

\begin{enumerate}
    \item We observe and identify the negative impacts of LDDs on the pipeline of the clustering-based topic model in a cross-lingual topic identification task.
    \item We introduce a dimension refinement component, implemented by either u-SVD or SVD-LR, into the current pipeline of the clustering-based topic model, which enables it to identify topics across languages.
    \item Our updated pipeline of the clustering-based topic model is shown to outperform the other state-of-the-art CLTMs on three datasets.
\end{enumerate}

\section{Methodology}

\subsection{Background: Pipeline of  Clustering-based Topic Model} \label{sec: pipeline of clustering-based topic model}

The pipeline of clustering-based topic model~\cite{grootendorstBERTopicNeuralTopic2022,zhangNeuralTopicModelling2022} contains four steps: Document Embedding Generation $\rightarrow$ Dimension Reduction $\rightarrow$ Document Clustering $\rightarrow$ Cluster Summarization. The first step adopts a pre-trained language model to embed documents into contextualized representations. The next step, Dimension Reduction, reduces the dimension of the representations for speeding up the subsequent clustering process. The Document Clustering step applies some clustering techniques, e.g., K-Means~\cite{zhangNeuralTopicModelling2022}, to the reduced representations for topic cluster identification. The last step, Cluster Summarization, reconstructs topic-word distribution by using word importance ranking metric, e.g., c-TF-IDF~\cite{grootendorstBERTopicNeuralTopic2022}, on each topic cluster. c-TF-IDF calculate the importance of the word $w$ in the cluster $k$ by

\begin{equation} \label{eq. c-tf-idf}
    \text{tf}_{w,k} \times \text{log}(1+\frac{A}{f_w}),
\end{equation}

\noindent where $\text{tf}_{w,k}$ is the word frequency of $w$ in the document cluster $k$, $A$ is the average word frequency of all clusters, and $f_w$ is the frequency of word $w$ across clusters. The higher value means the word $w$ is more representative to a cluster $k$.

\begin{figure*}
    \centering
    \includegraphics[width=2\columnwidth]{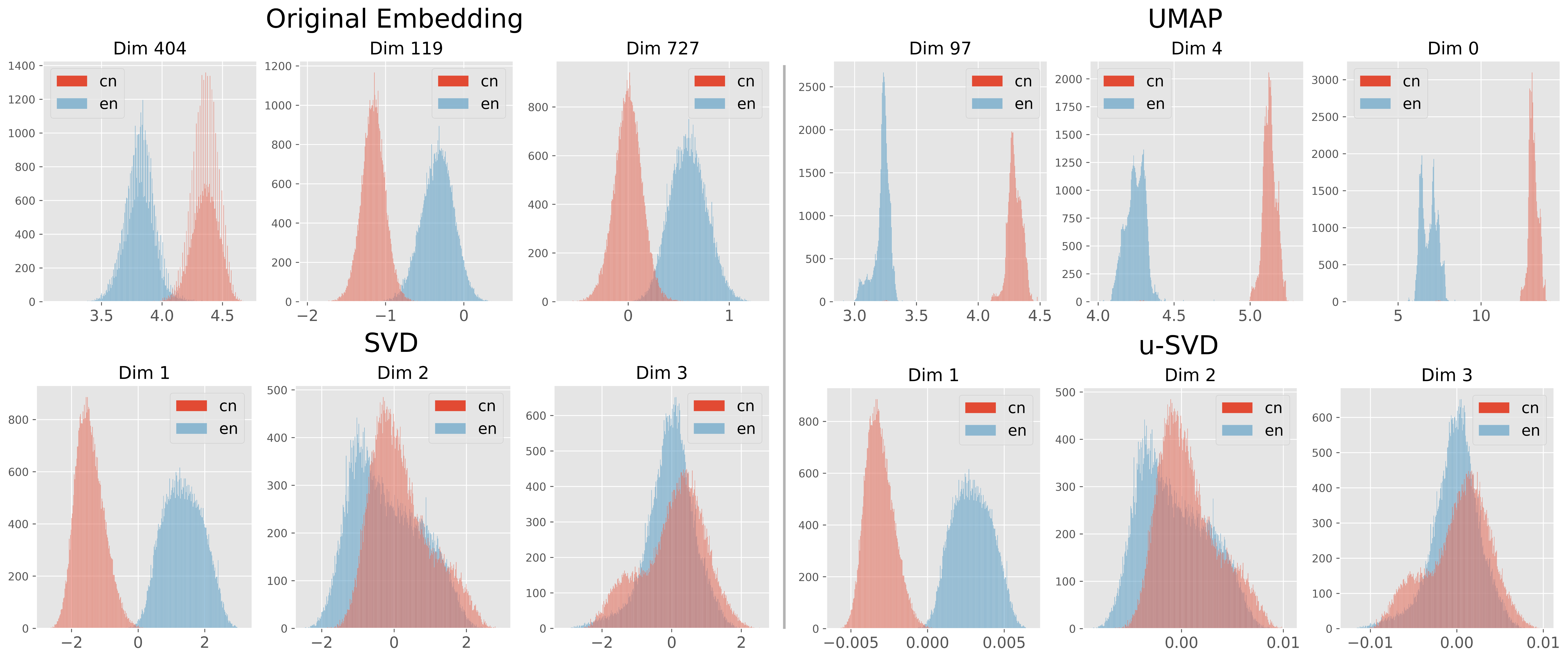}
    \caption{Top 3 language-dependent dimensions, sorted by t-statistic values, for original embeddings and embeddings reduced using UMAP, SVD and u-SVD. We utilize the Cohere multilingual model (see Section \ref{sec: multilingual language model}) to encode the documents in one of our experimental datasets, namely ECNews. The value distributions for Chinese (cn) and English (en) documents are indicated by red and blue, respectively. All UMAP, SVD, and u-SVD reduced the dimension size of the original representations from 768 to 100. Appendix~\ref{appendix:Rakuten_t_test} presents the same analysis to the other dataset, namely Rakuten Amazon.}
    \label{fig:distribution}
\end{figure*}

\subsection{Pipeline Adaption for Cross-lingual Topic Identification}

To adapt the current pipeline for cross-lingual topic identification, MLMs, such as Distilled XLM-R \citep{reimersMakingMonolingualSentence2020,conneauUnsupervisedCrosslingualRepresentation2020} and Cohere multilingual model, can be used in step 1 for embedding documents into language-agnostic representations~$E\in{R^{m\times{d}}}$, where $m$ is number of documents and $d$ is dimension of representations. However, we observe that a number of dimensions of MLMs’ representations retain language information. These dimensions are denoted as language-dependent dimensions (LDDs). To illustrate, we group documents written in language $l \in \{l_1, l_2\}$ and look into their representations. Let $e^{l}_i \in R^{m^l \times 1}$ be the values of $i$'th dimension for $m^l$ documents written in $l$. We compare the values of each dimension~$i \in d$ across two languages~$l_1$ and $l_2$ by performing a two-sample t-test on $e^{l_1}_i$ and $e^{l_2}_i$. We then sort all dimensions based on the corresponding t-statistics in descending order. As the larger t-statistic indicates the larger mean value difference across languages, we hereby identify LDDs. As shown in the upper-left subplot of Fig.~\ref{fig:distribution}, the original MLM embeddings show notable distinctions for documents written in two different languages, suggesting the presence of LDDs within the original embeddings. Furthermore, after applying UMAP, a dimension reduction approach used by previous cluster-based topic models~\citep{grootendorstBERTopicNeuralTopic2022,zhangNeuralTopicModelling2022}, even more significant LDDs are present (see the upper-right subplot of Fig.~\ref{fig:distribution}). This is likely to occur as UMAP focuses on capturing the local structure~\citep{mcinnesUMAPUniformManifold2020}.

LDDs adversely affect the subsequent document clustering process, as they disproportionately influence the distance calculations between documents during clustering. As a result, LDDs cause the algorithm to cluster documents by language rather than by their semantic meaning. In order to mitigate the negative impacts of LDDs, we repurpose the step 2 of the pipeline from a dimension reduction component to a dimension refinement component. Our dimension refinement component incorporates SVD, leveraging its notable feature that the reduced dimensions are orthogonal to one another. Note that previous researches have long applied SVD for topic modeling~\cite{deerwesterIndexingLatentSemantic1990,crainDimensionalityReductionTopic2012}, yet its usage has been confined to monolingual topic modeling for decomposing the term-document matrix to capture the latent semantic structure. We propose a novel approach that applies SVD to neutralize LDDs from the representations generated by MLMs and further reduces the influence of languages. Owing to the orthogonal decomposition property of SVD, when one dimension retains language information, the remaining dimensions are more likely to capture other types of information. The lower-left subplot of Fig.~\ref{fig:distribution} demonstrates that SVD consolidates the scattered LDDs into a concentrated set of reduced dimensions. 

We explore two implementations of dimension refinement components, namely unscaled SVD (u-SVD) and SVD with Language dimension Removal (SVD-LR). Both u-SVD and SVD-LR methods follow the same decomposition manner as the standard SVD, which is represented by  $E = U\Sigma{}V^T$. However, unlike standard SVD, u-SVD only utilizes $U\in{R^{m\times{r}}}$ to represent $m$ documents in $r$ reduced dimensions. Since $U$ is an orthonormal matrix, u-SVD reduces the influence of LDDs by ensuring that each dimension has a unit length. For instance, the lower-right subplot shows that u-SVD represents the dimensions using smaller scale (see x-axis) compared to the SVD in the lower-left subplot. By reducing the scale of dimensions, u-SVD decreases the negative contributions of LDDs in the subsequent clustering. u-SVD is a conservative approach as it reconciles the effects of LDDs without removing any dimension. In contrast, SVD-LR is more aggressive by removing the most influential LDD after performing SVD. Specifically, we represent the documents using $U\Sigma \in {R^{m\times{r}}}$ and use the two-sample t-test to identify the most influential LDD $\hat{r}$, which has the largest difference in the mean values of two languages. Then, SVD-LR removes $\hat{r}$ from $U\Sigma$.

\begin{algorithm}[ht]
\caption{Updated Pipeline for Cross-lingual Clustering-based Topic Model} \label{algo: pipeline of the model}
\begin{algorithmic}[1]
  \REQUIRE MLM, corpus, number of reduced dimensions $r$, number of topics $K$
  \STATE Obtain $E$ by embedding the corpus using the assigned MLM
  \STATE $U, \Sigma, V^T = \text{SVD}(E, r)$ 
  \IF{u-SVD}
    \STATE $E^{*} = U$
  \ELSIF{SVD-LR}
    \STATE Identify the most influential LDD $\hat{r}$ using two-sample t-test
    \STATE Obtain $E^{*}$ by removing $\hat{r}$ from $U\Sigma$
  \ENDIF
  \STATE $C_1, C_2, ..., C_K = \text{Kmeans}(E^{*}, K)$
  \STATE $\phi_1, \phi_2, ..., \phi_K = \text{c-Tf-IDF}(C_1, C_2, ..., C_K)$
  \RETURN $\phi_1, \phi_2, ..., \phi_K$
\end{algorithmic}
\end{algorithm}

Algorithm~\ref{algo: pipeline of the model} presents the updated pipeline, which is detailed as follows: (1) in line 1, documents are embedded using the MLM to obtain document representations~$E$, (2) from line 2 to line 8, we perform the dimension refinement step~\footnote{We use the SVD implementation from Dask package \url{https://www.dask.org}.} using either u-SVD or SVD-LR to obtain refined document representations~$E^{*}$, (3) in line 9, Kmeans algorithm~\footnote{We use Kmeans implementation with default parameters from scikit-learn package \url{https://scikit-learn.org/}.} are applied on $E^{*}$ to group documents into $K$ topic clusters, and (4) in line 10, we summarize and reconstruct the topic-word distribution for each topic cluster using c-TF-IDF (Eq.~\ref{eq. c-tf-idf}).  

\section{Experimental Setup}

\subsection{Dataset}
We conduct experiments using three datasets: (1) \textbf{Airiti Thesis} which consists of 163,150 pairs of English and Chinese thesis abstracts \cite{changEmployingWordMover2020}. On average, each abstract contains 165 words. (2) \textbf{ECNews} comprises 50,000 Chinese news and 46,850 English news articles, with an average length of 11 words per article. (3) \textbf{Rakuten Amazon} is a compilation of 25,000 Japanese and 25,000 English product reviews, with an average of 27 words per review. ECNews and Rakuten Amazon were used in the previous research for cross-lingual topic evaluation~\citep{wuInfoCTMMutualInformation2023}. Considering that ECNews and Rakuten Amazon primarily contain short documents, we include Airiti Thesis in our experiments to evaluate the performance on identifying topics in longer documents.

\subsection{Multilingual Language Model} \label{sec: multilingual language model}

We evaluate our proposed methods and compare them with other methods using three different MLMs: (1) \textbf{mBERT} \citep{devlinBERTPretrainingDeep2019} has been investigated for its capability on cross-lingual classification tasks \citep{pires-etal-2019-multilingual}. We use transformers\footnote{\url{https://github.com/huggingface/transformers}} to load bert-base-multilingual-cased\footnote{\url{https://huggingface.co/bert-base-multilingual-cased}} and use output of special classification token ([CLS]) to get the mBERT embedding for a document. (2) \textbf{Distilled XLM-R} \citep{reimersMakingMonolingualSentence2020} is designed for embedding a paragraph and is based XLM-R \citep{conneauUnsupervisedCrosslingualRepresentation2020}, which is superior than mBERT in parallel sentence retrieval \citep{libovicky-etal-2020-language}. We use sentence-transformers\footnote{\url{https://www.sbert.net}} to access Distilled XLM-R (paraphrase-xlm-r-multilingual-v1). (3) \textbf{Cohere multilingual model} has shown its capabilities in various cross-lingual retrieval tasks \citep{kamallooEvaluatingEmbeddingAPIs2023}. We use the Cohere multilingual model (embed-multilingual-v2.0) by the API\footnote{\url{https://txt.cohere.com/multilingual/}}.

\subsection{Baseline \& Competitor}

We compare three alternative baselines to show the effectiveness of using u-SVD and SVD-LR as dimension refinement step: (1) \textbf{original embedding}, referred as \emph{OE}, which is simply generated from the given MLM, (2) \textbf{UMAP}~\footnote{We use the implementation from umap-learn package \url{https://github.com/lmcinnes/umap}.}, which is the popular dimension reduction method, whose effectiveness in identifying monolingual topics has been shown (i.e., CETopic) \citep{zhangNeuralTopicModelling2022}, and (3) \textbf{pure SVD}, which is used as a benchmark to compare against u-SVD and SVD-LR. Moreover, we compare two recent cross-lingual topic models: (1) \textbf{Cb-CLTM} \cite{changWordEmbeddingbasedApproach2021} incorporates a cross-lingual word space into the generative process of latent Dirichlet allocation \citep{bleiLatentDirichletAllocation2003}. Cb-CLTM demonstrates its superior performances compared to other probabilistic cross-lingual topic models. To enable the Cb-CLTM, we use pre-aligned English-Chinese and English-Japanese word spaces from MUSE project\footnote{\url{https://github.com/facebookresearch/MUSE}}. (2) \textbf{InfoCTM} \cite{wuInfoCTMMutualInformation2023} is a neural topic model that identifies topics across languages based on the guidance of the given bilingual dictionary. InfoCTM is the state-of-the-art neural cross-lingual topic model. We follow the report of the InfoCTM to use a Chinese-English dictionary from MDBG\footnote{\url{https://www.mdbg.net/chinese/dictionary?page=cc-cedict}} and Japanese-English dictionary from MUSE project to link topics across languages.

\subsection{Evaluation Metric}

We measure the generated topics using two metrics widely adopted in previous CLTMs: CNPMI and Diversity. For each topic $k \in K$, we select top-$N$ represented words for $l_1$ and $l_2$ languages, denoted as $\mathcal{W}^{l_1}_{k,N}$ and $\mathcal{W}^{l_2}_{k,N}$.

\textbf{CNPMI} \citep{haoEmpiricalStudyCrosslingual2020,changWordEmbeddingbasedApproach2021,wuInfoCTMMutualInformation2023} measures the coherence of generated topic words across languages: 

\begin{equation}
    -\frac{1}{N^2}\sum_{w_i \in \mathcal{W}^{l_1}_{k,N}, w_j \in \mathcal{W}^{l_2}_{k,N}} \frac{\text{log}\frac{Pr(w_i, w_j)}{Pr(w_i)Pr(w_j)}}{\text{log}Pr(w_i,w_j)},
\end{equation}

\noindent where $Pr(w_i,w_j)$ is the co-occurring probability of words $w_i$ and $w_j$ and $Pr(w_i)$ is the marginal probability of $w_i$. For Airiti Thesis, we estimate the probability using the comparable abstracts in the Airiti Thesis. For ECNews and Rakuten, we measure the probability using comparable Wikipedia corpus~\footnote{We use the implementation \url{https://github.com/BobXWu/CNPMI} from the authors of InfoCTM~\cite{wuInfoCTMMutualInformation2023}.}. The CNPMI ranges from $-1$ (least coherent) to $1$ (most coherent), and we report the average CNPMI scores across $K$ topics.

\textbf{Diversity} \citep{diengTopicModelingEmbedding2020} measures the uniqueness of generated topic words across $K$ topics:

\begin{equation}
    \frac{|\bigcup_{1 \leq k \leq K}\mathcal{W}^{l_1}_{k,N}| + |\bigcup_{1 \leq k \leq K}\mathcal{W}^{l_2}_{k,N}|}{K \times 2 \times N},
\end{equation}

\noindent which ranges between 0 (the least diversity) and 1 (the highest diversity). To combine the two aspects, we further compute \textbf{Topic Quality (TQ)} \citep{diengTopicModelingEmbedding2020} as the product of max(0, CNPMI) and Diversity, providing a cohesive measure for our analysis. Note that positive CNPMI contributes to TQ because NPMI measurement positively correlates with human interpretability \citep{lauMachineReadingTea2014}. The topic with negative CNPMI are considered to be uninterpretable. 

We evaluate top 15 words ($N=15$) of each topic for CNPMI and Diversity. For more robust comparison, we re-run every method five times using different seeds and report the average performance.

\section{Results \& Analysis}

\subsection{Performance of Cross-lingual Topic Model}

\begin{table*}[ht]
\caption{Comparison of topic quality for baselines, competitors, and our proposed methods.}
\label{tbl:phi}
\centering
\resizebox{0.85\textwidth}{!}{%
\begin{tabular}{@{}lccccccccc@{}}
\toprule
\textbf{Dataset} & \multicolumn{3}{c}{\textbf{Airiti}} & \multicolumn{3}{c}{\textbf{ECNews}} & \multicolumn{3}{c}{\textbf{Rakuten Amazon}} \\ \midrule
\textbf{Metric} & \textbf{CNPMI} & \textbf{Diversity} & \textbf{TQ} & \textbf{CNPMI} & \textbf{Diversity} & \textbf{TQ} & \textbf{CNPMI} & \textbf{Diversity} & \textbf{TQ} \\ \midrule
OE & -0.244 & 0.570 & 0.000 & 0.022 & 0.554 & 0.012 & 0.009 & 0.290 & 0.003 \\
UMAP & -0.202 & 0.572 & 0.000 & 0.019 & 0.598 & 0.011 & 0.003 & 0.265 & 0.001 \\
UMAP-norm & -0.207 & 0.585 & 0.000 & 0.019 & 0.613 & 0.012 & 0.003 & 0.264 & 0.001 \\
SVD & -0.251 & 0.564 & 0.000 & 0.026 & 0.567 & 0.015 & 0.009 & 0.282 & 0.003 \\ \midrule
Cb-CLTM & -0.145 & \textbf{0.941} & 0.000 & 0.021 & 0.774 & 0.016 & 0.008 & 0.699 & 0.006 \\
InfoCTM & -0.087 & 0.917 & 0.000 & 0.044 & \textbf{0.905} & 0.040 & 0.033 & \textbf{0.856} & \textbf{0.028} \\ \midrule
SVD-LR & \textbf{0.179} & 0.571 & \textbf{0.103} & \textbf{0.087} & 0.741 & 0.065 & 0.032 & 0.607 & 0.019 \\
u-SVD & 0.171 & 0.603 & \textbf{0.103} & 0.086 & 0.823 & \textbf{0.071} & \textbf{0.037} & 0.665 & 0.025 \\ \bottomrule
\end{tabular}
}
\end{table*}

Table~\ref{tbl:phi} shows the performance of different methods on three datasets. We adopt the following settings. Cohere multilingual model is chosen as the MLM, which embeds every document into 768 dimensional representations. All dimension reduction methods reduce the original embedding from 768 to 100 dimensions. The number of topics (clusters) is set to 50 because InfoCTM \citep{wuInfoCTMMutualInformation2023} reports performances on this number for both ECNews and Rakuten Amazon.

The results clearly indicate that incorporating a clustering-based topic model pipeline with three baseline embeddings, including original embedding, UMAP, and SVD, does not perform well in terms of CNPMI and Diversity. We also use feature-wise min-max normalization on UMAP, resulting in UMAP-norm. However, UMAP-norm does not enhance performance. Both Cb-CLTM and InfoCTM exhibit high diversity scores. However, when applied to the Airiti dataset, they generate topics with negative CNPMI scores, suggesting that their generated topics are difficult to be interpreted by human \citep{lauMachineReadingTea2014}. The pipelines with u-SVD and SVD-LR result in less diverse topics than Cb-CLTM and InfoCTM but have better CNPMI and TQ on the Airiti and ECNews datasets. Moreover, InfoCTM, SVD-LR, and u-SVD reach comparable CNPMI and TQ on the Rakuten Amazon dataset. These results suggest that u-SVD and SVD-LR can generalize to datasets of different lengths.

\subsection{Performance on Different MLMs}

\begin{table*}[ht]
\caption{Topic quality of using three different MLMs.}
\label{tbl:mlm}
\centering
\resizebox{0.85\textwidth}{!}{%
\begin{tabular}{l|lll|lll|lll}
\hline
\multirow{2}{*}{Method}  & \multicolumn{3}{c|}{\bf mBERT} & \multicolumn{3}{c|}{\bf Distilled XLM-R} & \multicolumn{3}{c}{\bf Cohere Multilingual Model} \\ \cline{2-10}
& CNPMI      & Diversity & TQ     & CNPMI      & Diversity &TQ     & CNPMI      & Diversity &TQ    \\ \hline
OE    & -0.122     & 0.478  &    0.000   & -0.211     & 0.600  &    0.000   & -0.244     & 0.570     &  0.000  \\
UMAP & -0.190     & 0.421    &  0.000   & -0.198     & 0.536   &   0.000   & -0.202     & 0.572     &  0.000  \\
SVD            & -0.117     & 0.476    &  0.000   & -0.208     & 0.580    &  0.000   & -0.251     & 0.564     &  0.000  \\ \hline
SVD-LR     & -0.149     & 0.492   &   0.000   & 0.172      & 0.527    &  0.091   & \textbf{0.179}      & 0.571   &   \textbf{0.103}   \\
u-SVD    & \textbf{0.001}      & \textbf{0.591}    &  \textbf{0.000}   & \textbf{0.182}      & \textbf{0.629}     &  \textbf{0.115}  & 0.171      & \textbf{0.603}     &  \textbf{0.103}  \\ \hline
\end{tabular}%
}
\end{table*}

To test the generalizability of u-SVD and SVD-LR, we evaluate and compare performances on three MLMs, namely mBERT, Distilled XLM-R, and Cohere Multilingual Model, on the Airiti Thesis. All three MLMs generate document embedding with 768 dimensions. To benchmark with the results shown in Table~\ref{tbl:phi}, each document embedding is also reduced or refined to 100 dimensions, and the number of topic clusters is set to 50.

Table~\ref{tbl:mlm} reveals that when using mBERT, both SVD-LR and u-SVD achieve only marginal improvement, if any, on topic quality compared to other three baselines. This may be attributed to limited cross-lingual capability of mBERT because it is the first generation MLM. On the other hand, with the document representations generated by more capable MLMs, namely Distilled XLM-R and Cohere Multilingual Model, SVD-LR and u-SVD consistently demonstrates their robust performances and generate topic clusters with better topic quality.

\subsection{Sensitivity Analysis on the Size of Reduced Embeddings}

\begin{figure} [ht]
    \centering
    \includegraphics[width=\columnwidth]{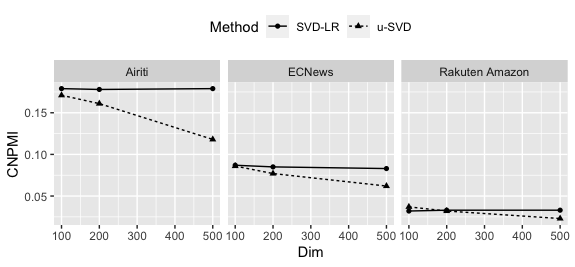}
    \caption{Sensitivity analysis of u-SVD and SVD-LR on different dimensions.}
    \label{fig:CNPMI Sensitivity}
\end{figure}

To better understand u-SVD and SVD-LR, we conduct sensitivity analysis on the size of embeddings. In this analysis, we use all three datasets and fix the number of cluster topics at 50. We reduce the document representations generated by Cohere Multilingual Model from 768 to 100, 200, and 500 to see their influence on the CNPMI.

Fig.~\ref{fig:CNPMI Sensitivity} shows that SVD-LR has a more robust result across different embedding dimensions. SVD-LR preserves the importance weight (i.e., $\Sigma$) of each dimension except for the most influential LDD, resulting in robust performance across various dimensions. On the contrary, u-SVD abandons the importance weight of dimensions from SVD to lessen the effect of LDDs. Thus, u-SVD is affected by those dimensions that originally had small singular values, leading to poorer outcomes when more dimensions are utilized. In summary, while both u-SVD and SVD-LR lose some information due to the elimination of LDDs, SVD-LR seems to lose fewer information when more dimensions are introduced.

\subsection{Qualitative Result}

\begin{table*} [ht]
\caption{Top representative words of 10 sampled topics from updated pipeline with u-SVD and SVD-LR}
\label{tbl:u-SVD_topic_words}
\centering
\resizebox{0.85\textwidth}{!}{
  \begin{tabular}{cl}
    \toprule
    \textbf{u-SVD} & \\
    \midrule
    Topic\#2& optical, 光學(optics), 雷射(laser), 發光(glow), laser, light, 元件(component),\\ 
     & led, 我們(we), 結構(structure)\\
    Topic\#7& 影像(image), image, 我們(we), 演算法(algorithm), 方法(method), propose,\\
     & algorithm, 提出(propose), method, video\\
    Topic\#9& 網路(network), 無線(wireless), 傳輸(transmission), 通訊(communication),\\
     & network, 我們(we), 使用(use), 系統(system), 提出(propose), propose\\
    Topic\#20& polymer, 高分子(polymer), 材料(material), surface, film, increase,\\
     & high, property, 結構(structure)\\
    Topic\#21& 投資(investment), 市場(market), 報酬(return), market, return, 股票(stock),\\
     & 交易(transaction), 指數(index), stock, 投資人(investor)\\
  \midrule
  \textbf{SVD-LR} & \\
  \midrule
    Topic\#1& optical, 發光(glow), 光學(optics), 雷射(laser), 元件(component), led, laser,\\
     & light, 結構(structure), 我們(we)\\
    Topic\#6& 影像(image), image, 我們(we), 演算法(algorithm), 方法, propose,\\
     & algorithm, 提出(propose), method, video\\
    Topic\#12& polymer, 高分子(polymer), 材料(material), surface, 表面(surface), film,\\
     & 結構(structure), increase, high, material\\
    Topic\#17& 網路(network), 無線(wireless), network, 傳輸(transmission), 我們(we),\\
     & 使用(use), 節點(node), 通訊(communication), 提出(propose), 服務(service)\\
    Topic\#21& 投資(investment), 市場(market), market, 報酬(return), return, 指數(index),\\
     & 交易(transaction), 股票(stock), stock, investor\\
  \bottomrule
\end{tabular}
}
\end{table*}

We apply the Cohere multilingual model to embed the Airiti dataset and use BERTopic~\citep{grootendorstBERTopicNeuralTopic2022}, which implements the previous pipeline of clustering-based topic model. Table~\ref{tbl:topic_words} shows the representative words for ten manually sampled topics generated by BERTopic. Each topic consists of top words purely from a single language and is misaligned by the semantic meaning. For instance, topics \#1 \& \#2 discuss the same topic but are separated into two topics. Table~\ref{tbl:u-SVD_topic_words} uses the same setting as Table~\ref{tbl:topic_words} but apply u-SVD and SVD-LR for dimension refinement. Most topics contain representative words across languages and are grouped by the semantic meanings of topics. For example, the concept of "Financial Market" is separated into two topics in Table~\ref{tbl:topic_words}, namely topics \#5 \& \#46, based on languages. On the contrary, as shown in Table~\ref{tbl:u-SVD_topic_words}, topic \#21 from u-SVD and topic \#21 from SVD-LR include the words of different languages yet with similar concept.

\section{Related Work}

\subsection{Clustering-based Topic Model} 

Recent works \citep{siaTiredTopicModels2020,zhangNeuralTopicModelling2022,grootendorstBERTopicNeuralTopic2022} have explored methods that cluster contextualized representations to identify topics from a corpus. \citet{siaTiredTopicModels2020} used the BERT model to encode each token into a representation, averaging these representations to obtain a document-level representation. They then applied K-means clustering to these document representations and reconstructed the topic-word distributions using a tf-idf weighting scheme. The coherence performance of their resultant topics was comparable to that of the traditional topic model, LDA \citep{bleiLatentDirichletAllocation2003}. Similarly, \citet{zhangNeuralTopicModelling2022} and \citet{grootendorstBERTopicNeuralTopic2022} proposed a pipeline consisting of four steps. First, they used language models, such as sentence BERT (SBERT), to encode documents into representations. Next, they applied the dimension reduction technique UMAP to these representations. In the third step, they used K-means clustering on the reduced representations to generate document clusters, each considered a topic cluster. Finally, they employed a word importance ranking method, c-Tf-IDF, to identify representative topic words. Their pipelines outperformed neural topic models in terms of both efficiency and topic quality. However, the proposed pipeline hasn't been evaluated in cross-lingual settings. Our study aims to fill this gap.

\subsection{Language-dependent Component} 

Several studies \citep{libovicky-etal-2020-language,zhaoInducingLanguageAgnosticMultilingual2021,changWordEmbeddingbasedApproach2021} have shown that MLM-generated representations contain language-dependent components (LDDs), which signal language identity and hinder cross-lingual transfer. To mitigate such LDDs, \citet{libovicky-etal-2020-language} noted that representations of the same language are closely located in the space. They recommend removing the language-specific mean from the mBERT representations as a solution. However, even after this adjustment, the resulting representations can still be utilized as features to predict the language accurately, suggesting that simply removing the language-specific means from the representations is insufficient. \citet{zhaoInducingLanguageAgnosticMultilingual2021} propose a method that requires parallel corpus to fine-tune mBERT and XLM-R for generating language-agnostic representations. The method fine-tunes the language model to align the sentence pairs from the parallel corpus. To further close the gap between languages, the method also constrains the representations of different languages to be distributed with zero mean and unit variance. Such an idea is close to our proposed u-SVD; however, u-SVD is a more efficient and appropriate method for models with ample parameters because it does not require parallel corpus and fine-tuning. \citet{changWordEmbeddingbasedApproach2021} observed that LDDs prevent their topic model from identifying topics across languages. They proposed training a logistic regression to identify the contributed dimensions (i.e., LDDs) for language identity and removed them from the representations. They found that removing the LDDs helped identify more cross-lingual topics. However, removing the LDDs directly from the original representations comes with the cost of losing semantic completeness. Our SVD-LR eases this issue because utilizing SVD helps us to consolidate the scattered language-dependent dimensions into one specific dimension. Therefore, SVD-LR only removes the most contributed LDD, potentially minimizing the risk of losing other semantic meanings. 

\section{Conclusion}
We investigate the problem with the current pipeline of clustering-based topic model when applied on multilingual corpus, which is caused by language-dependent dimensions in the multilingual contextualized embedding. To solve this problem, we propose two methods for dimension refinement, namely u-SVD and SVD-LR. Our experiments suggest that the updated pipeline with our proposed refinement component is effective in cross-lingual topic identification and results in more coherent topics than existing cross-lingual topic models.

\section*{Limitations}

This study only evaluates proposed dimension refinement components, u-SVD and SVD-LR, on three MLMs, namely mBERT, XLM-R, and Cohere Multilingual Model. We chose these three MLMs because of their extensive investigations in cross-lingual retrieval tasks. The future work may investigate more other MLMs such as LASER\footnote{\url{https://github.com/facebookresearch/LASER}}, Universal Sentence Encoder\footnote{\url{https://www.kaggle.com/models/google/universal-sentence-encoder/}}, and OpenAI embedding API \footnote{\url{https://platform.openai.com/docs/api-reference/embeddings}}. Extensive experiments on more language pairs are another future work since we only evaluate two English-Chinese datasets and one English-Japanese dataset. It is worth noting that our proposed methods are effective in language pairs from distant and different language families. Furthermore, it's also crucial to investigate our methods for datasets with more than two languages, such as EuroParl. 




\end{CJK*}

\bibliography{references}

\appendix

\section{Two-sample t-test on MLM embeddings for Rakuten Amazon dataset}
\label{appendix:Rakuten_t_test}
\begin{figure*}
    \centering
    \includegraphics[width=2\columnwidth]{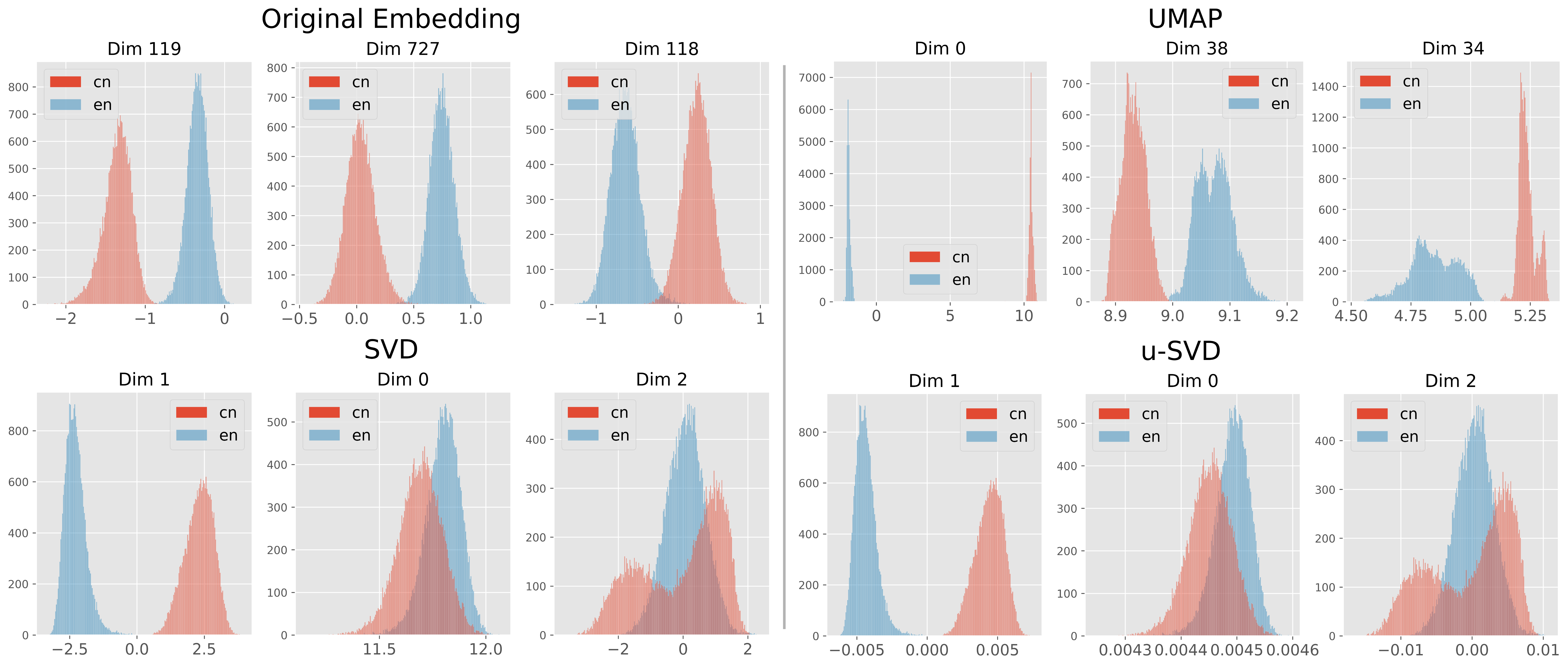}
    \caption{Top 3 language-dependent dimensions, sorted by t-statistic values, for original embeddings and embeddings reduced using UMAP, SVD and u-SVD on Rakuten Amazon dataset.}
    \label{fig:Rakuten_distribution}
\end{figure*}

We use the same setting as in Fig~\ref{fig:distribution} to display the top three language-dependent dimensions of the Rakuten Amazon dataset in Fig~\ref{fig:Rakuten_distribution}.

\end{document}